\documentclass[runningheads]{llncs}
\usepackage[T1]{fontenc}
\usepackage{amsmath}
\usepackage{amssymb}
\usepackage{siunitx}
\usepackage{pdfpages}
\usepackage{caption}
\usepackage{subcaption}
\usepackage{layouts}
\usepackage[export]{adjustbox}
\PassOptionsToPackage{hyphens}{url}\usepackage{hyperref}
\usepackage{graphicx}
\begin{document}
\sloppy
\title{Transient Hemodynamics Prediction Using an Efficient Octree-Based Deep Learning Model}
\titlerunning{Efficient Transient Hemodynamics Prediction}
\author{Noah Maul\inst{1,2} \and
Katharina Zinn \inst{1,2} \and
Fabian Wagner \inst{1} \and
Mareike Thies \inst{1} \and
Maximilian Rohleder \inst{1,2} \and
Laura Pfaff \inst{1,2} \and
Markus Kowarschik  \inst{2} \and
Annette Birkhold \inst{2}\and
Andreas Maier \inst{1}
}
\authorrunning{N. Maul et al.}
\institute{Pattern Recognition Lab, FAU Erlangen-Nürnberg, Germany \and
Siemens Healthcare GmbH, Forchheim, Germany\\
\email{noah.maul@fau.de}}
\maketitle              

\begin{abstract}
Patient-specific hemodynamics assessment could support diagnosis and treatment of neurovascular diseases. 
Currently, conventional medical imaging modalities are not able to accurately acquire high-resolution hemodynamic information that would be required to assess complex neurovascular pathologies.
Therefore, computational fluid dynamics (CFD) simulations can be applied to tomographic reconstructions to obtain clinically relevant information.
However, three-dimensional (3D) CFD simulations require enormous computational resources and simulation-related expert knowledge that are usually not available in clinical environments.
Recently, deep-learning-based methods have been proposed as CFD surrogates to improve computational efficiency.
Nevertheless, the prediction of high-resolution transient CFD simulations for complex vascular geometries poses a challenge to conventional deep learning models. 
In this work, we present an architecture that is tailored to predict high-resolution (spatial and temporal) velocity fields for complex synthetic vascular geometries. 
For this, an octree-based spatial discretization is combined with an implicit neural function representation to efficiently handle the prediction of the 3D velocity field for each time step. 
The presented method is evaluated for the task of cerebral hemodynamics prediction before and during the injection of contrast agent in the internal carotid artery (ICA). 
Compared to CFD simulations, the velocity field can be estimated with a mean absolute error of \SI{0.024}{\meter\per\second}, whereas the run time reduces from several hours on a high-performance cluster to a few seconds on a consumer graphical processing unit.

\end{abstract}
\section{Introduction}
Vascular diseases are globally the leading cause of death  \cite{who_cvd_2011} and thus optimal medical diagnosis and treatment are desirable.
A detailed understanding of vascular abnormalities is essential for treatment planning, which includes hemodynamic information such as blood velocity and pressure. However, especially for neurovascular pathologies  quantitative blood flow information at sufficiently high resolution is difficult to acquire by conventional medical imaging alone. Hence, measurements, e.g., three-dimensional (3D) tomographic reconstructions of the abnormality, are usually coupled with computational methods. 
These methods include 3D computational fluid dynamics (CFD) simulations, that are based on solving partial differential equations (PDEs) numerically. However, the setup of CFD simulations requires domain-specific knowledge and enormous computational resources, which are usually not available in a clinical environment. Even without the required preprocessing steps, simulations require several hours of runtime on high-performance computing clusters.
Recently, deep-learning-based models have been proposed to approximate CFD results with different input data, acting as surrogate models. Once trained, these models allow to reduce runtime during inference \cite{Feiger2020,yuan_2022,taebi_2022,edward_2020,rutkowski_2021,du_2022}. 
However, the presented approaches either regress a desired hemodynamic quantity directly (without estimating the whole 3D velocity or pressure field), work on small volumetric patches (only possible with suitable input data, e.g., magnetic resonance tomography), or are limited to steady-state simulations. 
To the best of our knowledge, there exists no prior work on high-resolution transient CFD surrogate models for complex vascular geometries.
We hypothesize that lacking suitable deep learning architectures and associated data representations for time-resolved volumetric vascular data so far prohibit more accurate methods.
Particularly, uniformly spaced voxel volumes are inherently ill-suited for discretizing neurovascular geometries, as the size of the smallest feature defines required resolution and voxel size. Further, the resulting volumes are usually sparse, meaning that a large part of the volume is not covered by (image-able) vascular structures and is, therefore, not contributing to clinically relevant hemodynamics. 
Convolutional neural networks (CNNs) are in principle well-suited for flow prediction as they exploit local coherence \cite{xie_2018}. However, this would require huge computational resources due to the sparse data.
Even more challenging is the application of CNNs to transient simulations, where the time dimension must be considered in the architecture, leading to further increase of computational complexity.

In this work, we present a computationally efficient deep learning architecture that is designed to infer high-resolution velocity fields for transient flow simulations in complex 3D vascular geometries. 
We utilize octree-based CNNs \cite{wang_2017,wang_2020}, enabling sparse convolutions, to efficiently discretize and process neurovascular geometries with a high spatial resolution. Furthermore, instead of utilizing four-dimensional or auto-regressive architectures, we formulate the regression task as an operator learning problem. 
This concept has been introduced to solve PDEs in general \cite{Lu2021,li2020fourier} and has also been applied to medical problems, such as tumor ablation planning \cite{meister_2022}. 
The presented method is evaluated for the task of cerebral hemodynamics prediction based on a 3D digital subtraction angiography (3D DSA) acquisition, which is usually performed for treatment planning of neurovascular procedures.

\section{Methods}
\subsection{Problem Description and Method Overview}
In this work, the goal is to train a deep-learning-based CFD surrogate model to approximate the velocity field given the vascular geometry, and the associated boundary and initial conditions, which are the commonly used input data for a hemodynamics CFD simulation.
We aim to learn the solution operator for the incompressible Navier-Stokes equations
\begin{equation}
\begin{split}
\frac{\partial \mathbf{u}}{\partial t}
    + \mathbf{u} \cdot \nabla \mathbf{u}
    &= - \nabla p + \nu \nabla^2 \mathbf{u}  \\
    \nabla \cdot \mathbf{u} &= 0 \, ,
\end{split} 
\end{equation}
where $\mathbf{u}$ denotes the velocity, $p$ the pressure and $\nu$ the kinematic viscosity of a fluid. 
For this, the following steps are carried out. First, synthetic cerebral vessel trees are generated and preprocessed for simulation. Second, boundary conditions (BCs) are chosen from a representative set and a CFD solver is employed to calculate a set of reference solutions that describe the blood and total flow during injection of a contrast agent (CA). 
Third, the reference solutions are postprocessed, resulting in a dataset that is used to train a neural network. After training, the  surrogate model is applied to unseen data. 

\subsection{Physics Model of 3D DSA Acquisition}
Our physics model is designed to describe the physiological blood flow and the effect of the injection of contrast agent (CA) during a neurovascular 3D  DSA acquisition. 
As CA is commonly injected into the ICA, we focus our analysis on this case. Like previous studies \cite{Sun_2012,Waechter_2008}, we assume that the density difference between CA and blood is negligible and model the mixture as a single-phase flow. 
However, due to resistances downstream, the blood flow rate before injection $Q_\text{B}$ and the injection flow rate $Q_\text{CA}$ do not simply add up but can be described with a mixing factor $m$ \cite{Sun_2012,Mulder_2011},
$Q_\text{T}(t) = Q_\text{B}(t) + m \cdot Q_\text{CA}(t) \, ,$
where $Q_\text{T}$ denotes the total flow rate.
As in previous work \cite{Sun_2012}, a mixing factor of $0.3$ is chosen. Further, we model the compliance and resistance of the contrast flow through the catheter by an analogous electrical network consisting of a resistor and a capacitor \cite{Waechter_2008}. 
Therefore, the injection flow rate $Q_\text{CA}(t)$ can be described by 
\begin{equation}
    Q_\text{CA}(t) =
    \begin{cases}
    0  & t < T_\text{S} \\
    Q_\text{CA}^{\text{max}} \cdot \left(1 - \text{e}^{-(t-T_\text{S})/T_\text{L}}\right) & t \geq T_\text{S} 
    \end{cases} \, , 
\end{equation}
where $T_\text{S}$ refers to the injection start time, $T_\text{L}$ to the time of the lag and $Q_\text{CA}^{\text{max}}$ to the maximum injection rate  that is set to \SI{2.5}{\milli \liter \per \second}.
Moreover, the physiological blood flow rate should reflect real conditions and is therefore derived from reported values in literature. 
To generate a set of representative inflow conditions of humans, we follow the approach of Ford et al. \cite{Ford_2005} and Hoi et al. \cite{Hoi_2010}. 
They showed how the inflow waveform for the ICA can be modeled by the mean flow rate, cardiac cycle length and age. 
We select a set of flow rates~$\in \{3.4,4.4,5.4\}\,\si{\milli\liter\per\second}$ and  cardiac cycle lengths~$\in \{785,885, 985\}\,\si{\milli\second}$ that correspond approximately to the mean $\pm$ one standard deviation. 
For each combination of the sets, two waveforms (young and elderly) are generated, leading to $18$ different inflow waveforms in total. 
The flow is modeled as laminar and blood as a Newtonian fluid with a kinematic viscosity $\nu$ of \SI{3.2e-6}{\meter\squared \per \second} and a density of \SI{1.06e3}{\kilogram\per\meter\cubed}. Vessel walls are assumed to be rigid,  no-slip and zero-gradient pressure BCs are applied. 
The outlet BCs are set according to the flow-splitting method \cite{Chnafa_2018}, which avoids unrealistic zero pressure outlet BCs. 
The algorithm starts at the inlet (most distal part of the ICA) and determines the flow split ratio between the branches at each bifurcation. When an outlet is reached, the associated flow rate is assigned as the BC. 

\subsection{Generation of Synthetic Vascular Geometries}
\begin{figure}
    \centering
    \includegraphics[width=.65\linewidth]{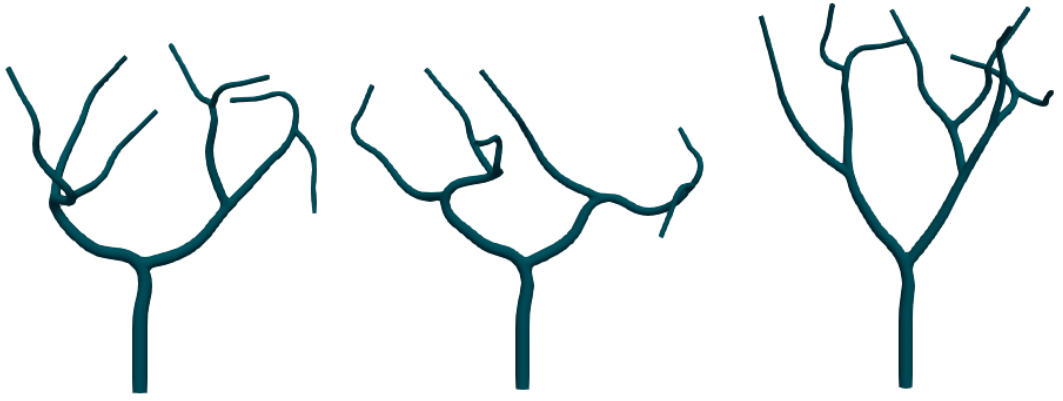}
    \caption{Three synthetic vascular trees generated for the dataset.}
    \label{fig:trees}
\end{figure}

The surrogate model is trained and evaluated with a dataset that comprises synthetic neurovascular geometries in combination with physiological BCs.
Surface meshes of vessel trees are automatically generated using the 3D modeling software \textit{Blender} (Blender, version $2.92$, Blender Foundation) and the \textit{Sapling Tree Gen} addon (Sapling Tree Gen, version 0.3.4, Hale et al.) that is based on a sampling algorithm by Weber et al. \cite{sapling_paper}.
The root branch is modeled as a synthetic ICA, where the radius is chosen uniformly random between $1.62$ and \SI{1.98}{\milli\meter} \cite{Chnafa_2017}. 
To cover a large variety of bifurcation types, bifurcation angles are chosen uniformly distributed between $35$ and \SI{135}{\degree} with an additional vertical attraction factor of $2.6$ \cite{sapling_paper}. 
To ensure a developed flow and avoid backflow at the outlets, flow extensions with an approximate length of five times the respective vessel diameter are added to the inlet and all outlets. Three synthetic trees are depicted in Fig~\ref{fig:trees}. 
Centerlines are calculated on the resulting surface mesh and a locally radius-adaptive polyhedral volumetric mesh containing five prismatic boundary layers to capture steep gradients near the vessel wall is generated \cite{Vmtk_2018,Weller_1998}. 

\subsection{CFD Simulation Setup}
\label{sec:simulation}
For each vessel tree, the flow rate, cardiac cycle length and the age is sampled uniformly random and the resulting inflow curve is set as the inflow BC (plug flow velocity profile) for the simulation. 
The CFD software \textit{OpenFOAM} (OpenFOAM, version 8, The OpenFOAM Foundation) \cite{Weller_1998} is employed and 
second order schemes are chosen for space and time discretization. An adaptive implicit time-stepping method with a maximum timestep of \SI{1}{\milli\second} is employed. Overall, four cardiac cycles are simulated. The first one is used to wash out initial transient effects, while the second one reflects the hemodynamics before contrast injection. 
The virtual injection of CA starts with the third cardiac cycle and is continued until the end of the simulation. 
Each simulation is executed utilizing 40 CPUs on a high-performance cluster requiring several hours of runtime.

\subsection{Deep Learning Architecture}

\begin{figure}
    \centering
    \includegraphics[width=\linewidth]{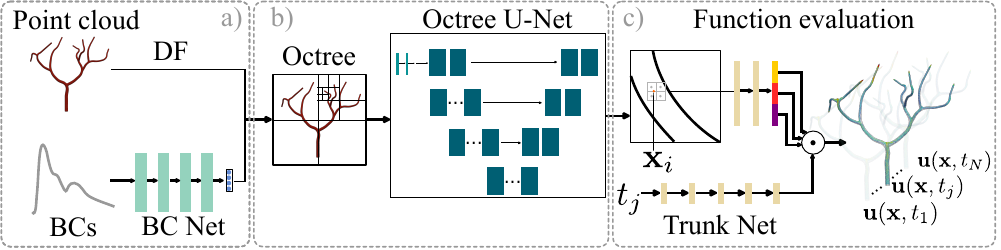}
    \caption{Overview of the proposed architecture that consists of three buildings blocks: geometry and BC encoding (a), octree construction and octree U-Net inference (b), and neural function evaluation (c).}
    \label{fig:architecture}
\end{figure}

The proposed deep learning model is designed to infer the velocity field from the geometry and BCs. The model consists of three main building blocks, as illustrated in Fig.~\ref{fig:architecture}.
In the first block (a), a 3D point cloud representation of the geometry and the BCs (inflow waveform) are processed to compute node features for each point. 
In the second block (b), an octree \cite{maegher_1982} is calculated from the point cloud, and passed to an octree-based neural network. 
In the third block (c), the output of the network is regarded as a continuous neural representation of the velocity field function that can be evaluated at a set of spatiotemporal points in parallel. The individual building blocks are elaborately presented in the following. 
\subsubsection{a) Geometry and Boundary Condition Processing}
The input to the model consists of a point cloud representation of the geometry and the BCs. In the dataset, the inflow waveform provides sufficient information to solve the Navier-Stokes equations as the outflow BCs are calculated from the inflow waveform and the geometry (flow-splitting method).
Also, the initial conditions are shared across all cases.
The distance field (DF) of the point cloud to the tree surface is calculated and the 1D inflow waveform is supplied to a 1D CNN (BC Net).
This network contains four blocks, each consisting of a convolution, pooling, and leaky rectified linear unit (LReLU) layer.
As a last operation, the output is pooled to a four-dimensional vector by averaging over the feature maps.
The feature vector and the DF are concatenated, repeating the same feature vector for all field points, and together form the point cloud features.

\subsubsection{b) Octree Construction and U-Net}
The octree data structure is built from the spatial information of the point cloud \cite{wang_2017} with a maximum depth level of ten, leading to an isotropic node resolution of \SI{0.15}{\milli\meter} on the finest level. 
At each level until maximum depth, a node is refined if it contains one or more points, such that the adaptive octree resolution can be controlled by the input point cloud. 
The point features that lie within a node on the finest level are averaged and considered the node features.
To learn the neural representation of the velocity field, a U-Net architecture \cite{ronneberger_2015} is employed. 
Wang et al. \cite{wang_2017} introduced a convolution layer that acts directly on the octree structure, such that calculations are only performed in spatial locations where necessary (inside vessels).
The U-Net consists of three downsampling (strided convolution) and three corresponding upsampling (transposed convolution) steps, starting at the maximum octree depth level down to the seventh level.
The four encoder levels consist of two, three, four, and six residual bottleneck blocks \cite{he_2016}, respectively. Each decoder level comprises two bottleneck residual blocks. A bottleneck residual block consists of two $3\times3$~convolutions and one $1\times1$~convolution for the residual connection. The LReLU is used as an activation function.

\subsubsection{c) Neural Function Evaluation}
The output of the U-Net is regarded as a continuous neural representation of the velocity field that can be evaluated at arbitrary spatiotemporal points.
Meister et al. \cite{meister_2022} employ an operator learning approach (mapping between function spaces) to predict a temperature distribution using the DeepONet architecture \cite{Lu2021} by approximating the temporal antiderivative operator applied on each voxel of a 3D grid. This allows to efficiently evaluate a large number of spatiotemporal points in parallel.
We extend this approach by exploiting the high-resolution octree representation. To evaluate the time-resolved velocity field at an arbitrary spatial point $\mathbf{x} \in \mathbb{R}^3$, the corresponding feature vector is calculated by trilinear interpolation between the neighboring octree node features at the finest octree level.
The feature vector is transformed by two fully connected layers and the result is split in three equally sized parts $\mathbf{b}_\mathbf{x} = (\mathbf{b}^{1}_\mathbf{x}, \mathbf{b}^{2}_\mathbf{x},\mathbf{b}^{3}_\mathbf{x})$, $\mathbf{b}^{i}_{\mathbf{x}} \in \mathbb{R}^{d}$, that contain the time dynamics information of the three approximated velocity vector components $\hat{\mathbf{u}}=(\hat{u}_1, \hat{u}_2, \hat{u}_3)$ at $\mathbf{x}$.
A series of five fully connected layers and LReLU activation functions (trunk net \cite{Lu2021}) receives the time point $t$ as an input and result in a latent vector $\mathbf{r_t} \in \mathbb{R}^d$.
To evaluate $\hat{\mathbf{u}}(\mathbf{x}, t)$, three dot products are calculated 
\begin{equation}
\label{eq:dot_prod}
   \hat{u}_i(\mathbf{x}, t) = \sum^{d}_{k = 1} b^{\,i}_{\mathbf{x}, k}\, r_{t,k} + c_i \, ,
\end{equation}
where $i \in \{1,2,3\}$ and $c_i$ denotes a learnable bias for each velocity component. The algorithmic steps are also illustrated in Fig.~\ref{fig:architecture}.
Note that this formulation only requires a single forward pass of the octree U-Net for each full time-resolved velocity field. Further, batches of $\{\mathbf{b}_\mathbf{x_i}\}$ and $\{\mathbf{r_{t_j}}\}$ can be calculated independently and in parallel.

\subsection{Training Setup}
To train and test the proposed model for the regression task, a suitable dataset is constructed using the synthetic geometries and simulated hemodynamics data. Overall, a virtual CA injection is simulated for 45 different virtual vessel trees, as described in Sec.~\ref{sec:simulation}. The velocity field is saved 30 times per second to be consistent with clinical 3D DSA image data that is usually acquired at 30 frames per second. Due to differing cardiac cycle lengths, this leads to a varying number of samples per simulation.
The samples are then split by geometry in training (35), validation (5) and test set (5). For each sample, the cell centers from the volumetric mesh are used as an input point cloud to construct the octree. The same locations are also utilized to evaluate the learned velocity field. As a data augmentation technique during training, the point clouds are randomly rotated and translated, such that a greater variety of octrees can be constructed.
To optimize the network parameters, the mean absolute error $\frac{1}{3N} \sum_{j=1}^N \sum_{i \in \{1,2,3\}} |\hat{u}_{ij} - u_{ij}|$ between predicted and actual velocity components is calculated for $N$ spatiotemporal points.
Ten time points are randomly chosen per simulation and batched to reduce training time. The ADAM optimizer \cite{Kingma_2015} is employed to train the network until convergence and the model with the lowest validation loss is selected for evaluation.

\section{Results}

The method is evaluated on the test set that contains the simulations of five separate vessel trees. Like in the training phase, the network is queried at the cell centers of the CFD mesh and at 30 time steps per second until the end of the third cardiac cycle. We quantitatively compare the predictions of our method with the CFD simulations.

\subsection{Quantitative Evaluation}
\begin{figure}
    \centering
    \begin{subfigure}{1.95in}
     \centering
     \includegraphics[width=\linewidth]{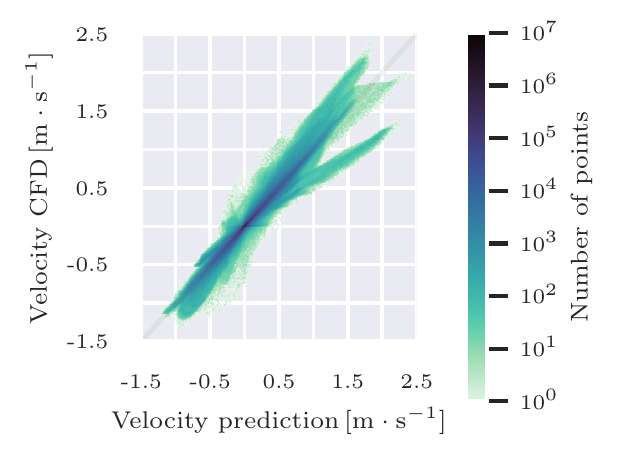}
    \end{subfigure}
    \begin{subfigure}{0.48\linewidth}
     \centering
     \includegraphics[width=\linewidth]{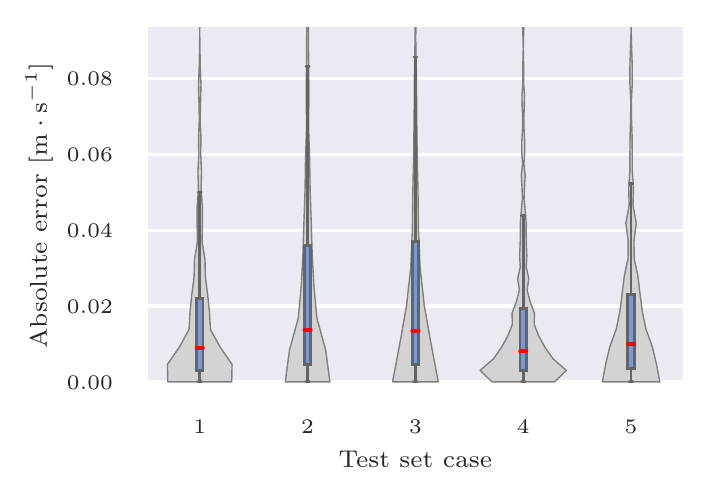}
    \end{subfigure}
    \caption{Statistical evaluation of the predicted velocity field for the test set cases. Due to the large number of points, only the time-averaged velocities are considered. 
    The left figure depicts the joint histogram of predicted and CFD (time-averaged) velocities (all components) over the entire test set (mean absolute error of \SI{0.023}{\meter\per\second} and $R^2$ of 0.97). Please note the logarithmic colormap. The right figure displays the error distribution for each individual test case. The y-axis limit is set to the maximum $90$-percentile of the cases. }
    \label{fig:statistics}
\end{figure}

The overall mean, standard deviation, and median of the absolute error across the whole test set is $0.024$, $0.043$ and \SI{0.010}{\meter\per\second}, respectively.
To avoid time-consuming statistical processing, the time-averaged velocity field per case is computed for the network prediction and the CFD simulation. 
In Fig. \ref{fig:statistics}, a regression plot over the time-averaged test set (left), as well as the error distribution for each individual case (right) is depicted. The mean absolute error of the time-averaged velocities is \SI{0.023}{\meter\per\second} and a coefficient of determination ($R^2$) of $0.97$ is determined for the prediction.
For some points, the network tends to underestimate the velocities. The quartiles of the error distribution show that \SI{75}{\percent} of all velocities can be estimated with an error smaller than \SI{0.040}{\meter\per\second}.

\subsection{Qualitative Evaluation}
\begin{figure}
    \centering
    \includegraphics[width=\linewidth]{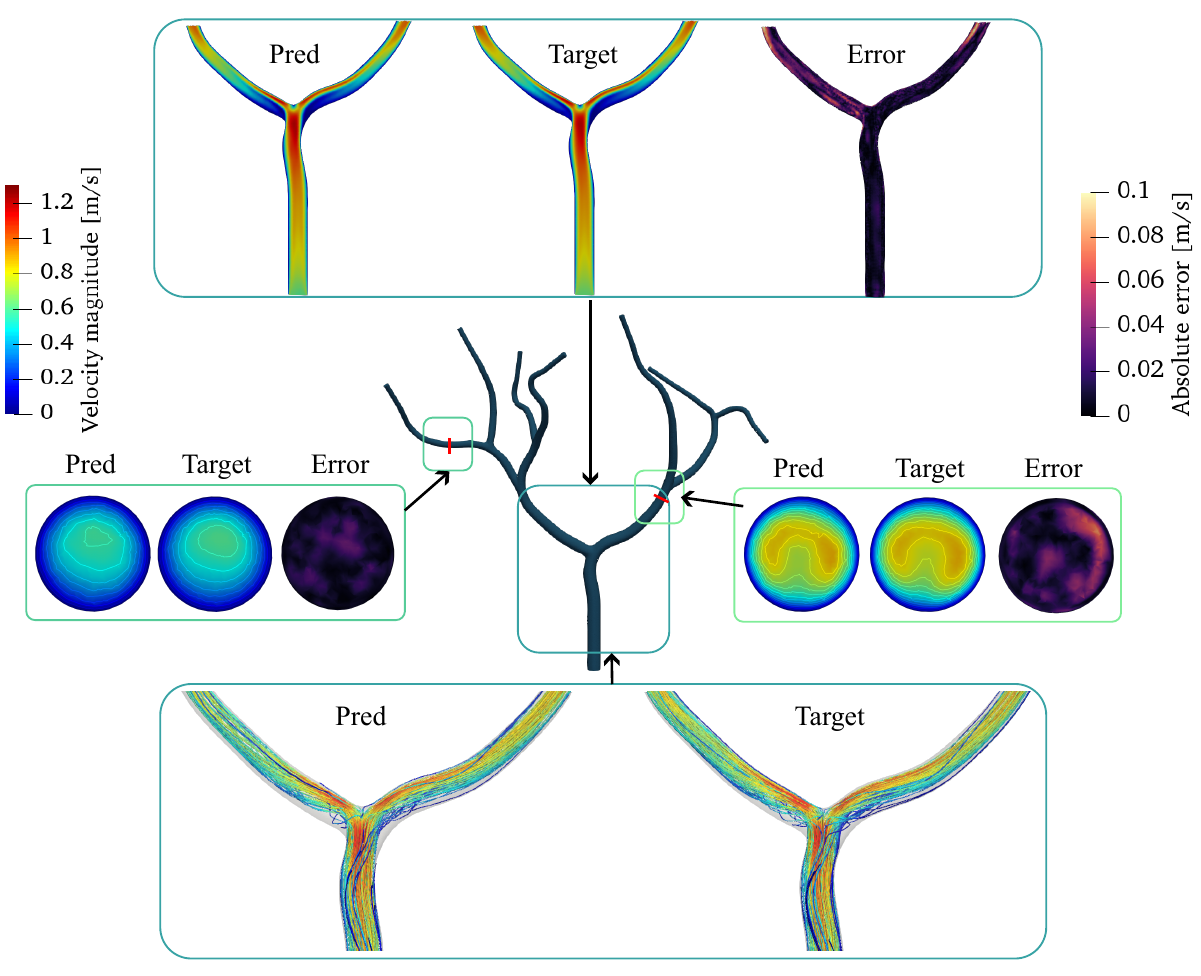}
    
    \caption{Evaluation of the (time-averaged) velocity field at three locations for test case five (median of mean absolute error across cases, visualized in Fig. \ref{fig:statistics}). All slices depict the velocity magnitude and the corresponding error between CFD and network prediction for a cross section of the 3D model. Streamline plots provide further information about the velocity field components by visualizing the trajectory of virtual particles.}
    \label{fig:qualitative}
\end{figure}
\begin{figure}
    \centering
    \includegraphics[width=\linewidth]{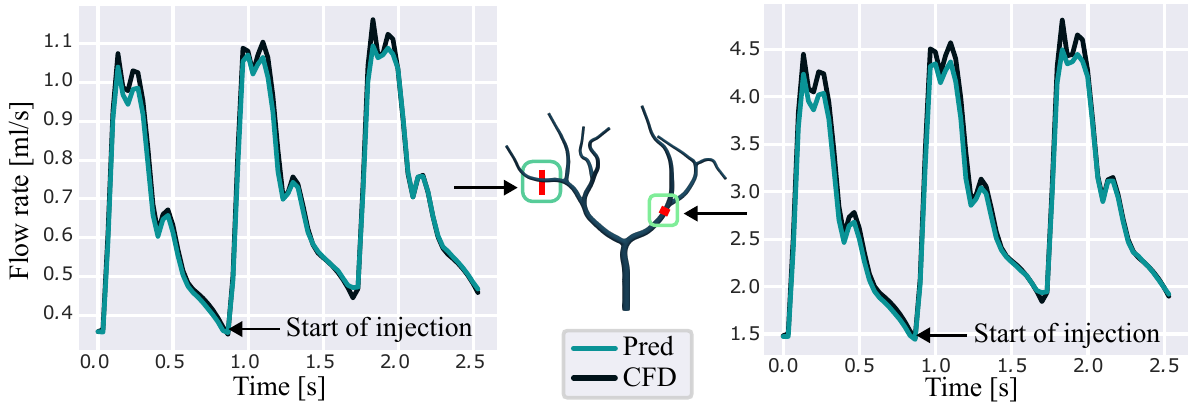}
    
    \caption{Comparison of volumetric flow rates  at two slices for test case five (median of mean absolute error across cases, visualized in Fig. \ref{fig:statistics})}
    \label{fig:qualitative_flowrate}
\end{figure}
We qualitatively evaluate the time-averaged velocity distributions for test case five (median of mean absolute error across cases), which is depicted in Fig.~\ref{fig:qualitative}. 
Three cross-sectional slices of the velocity magnitude at different locations and streamlines at the first bifurcation are depicted. The first slice is positioned far downstream after three bifurcations at a position where a less complex flow is expected (small radius and curvature, some distance from the previous bifurcation).
To evaluate more complex flow scenarios, an additional slice is placed right before the second bifurcation. The third slice shows the cross section of the first bifurcation.
Overall, both slice and streamline plots show reasonable agreement between predicted and actual velocity values as well as distributions. 
Additionally, the volumetric flow rate over time is compared for the first two slices, which is displayed in Fig~\ref{fig:qualitative_flowrate}. The flow rates agree reasonable, but are slightly underestimated for the three systoles (first and secondary peaks).

\subsection{Runtime Evaluation}
We evaluate the GPU-time of the model on a NVIDIA Quadro RTX 3000 ($6$ GB memory) graphical processing unit for one vessel tree from the test set. The overall time can be split in three parts: BC Net and U-Net forward pass $t_\text{net}$, spatial function evaluation $t_\text{spatial}$ and temporal $t_\text{temporal}$ function evaluation (trunk net forward pass and dot products). 
The spatial function evaluation comprises the octree interpolation and subsequent network transformation, whereas temporal evaluation refers to the execution of the trunk net and the dot products (visualized in Fig.~\ref{fig:architecture}). 
The spatial evaluation is performed with a batch of $10^6$ coordinates. For the temporal evaluation, the trunk net is executed with a batch of $100$ timesteps and the dot products are calculated between the output and the $10^6$ feature vectors.
We measure the runtimes $204.5\pm\SI{2.7}{\milli\second}$,  $92.2\pm\SI{4.9}{\milli\second}$ and $23.8\pm\SI{1.5}{\milli\second}$ for $t_\text{net}$, $t_\text{spatial}$ and $t_\text{temporal}$ respectively, where mean and standard deviation are calculated across $100$ runs for each part. Hence, assuming a velocity prediction for $N_\text{s}$ spatial points and $N_\text{t}$ temporal points, the overall runtime is approximately $t_\text{net} + N_\text{s} / 10^6 \cdot t_\text{spatial} +N_\text{t}/10^2 \cdot t_\text{temporal}$.

\section{Discussion}

Computational methods for 3D hemodynamics assessment of neurovascular pathologies require enormous computational resources that are usually not available in clinical environments. 
We presented a method that is tailored to predict the high-resolution (spatial and temporal) velocity field given a (complex) vascular geometry and corresponding boundary and initial conditions. By combining an explicit octree discretization with an implicit neural function representation, our proposed model enables the approximation of transient 3D hemodynamic simulations within seconds. 
We evaluated the method for the task of hemodynamics prediction during a 3D DSA acquisition for virtual cerebral vessels trees, where CA is injected into the ICA. Once trained, the velocity field can be inferred for unseen vascular geometries with a mean absolute velocity error of $0.024\,\pm\, \SI{0.043}{\meter\per\second}$.
Our quantitative and qualitative evaluation showed good agreement between the prediction of our model and the CFD ground truth. 
Existing approaches for predicting hemodynamics with machine learning surrogate models either regress a derived low-dimensional hemodynamic quantity directly (without outputting 3D velocity or pressure fields) \cite{Feiger2020}, rely on magnetic resonance imaging input \cite{edward_2020,rutkowski_2021}, or predict 3D steady-state simulations \cite{du_2022,Li_2021} with fixed BCs.
Compared to this, our method allows the prediction of high-resolution unsteady velocity fields for varying BCs. 
Raissi et al. \cite{raissi_2020} use physics-informed neural networks (PINNs) to predict high-resolution transient hemodynamics inside an aneurysm  assuming that the concentration field of a transported passive scalar can be measured. In contrast to our work, this allows to infer the underlying velocity field without knowledge of the BCs.
However, PINNs usually must be retrained in a self-supervised manner for each inference case and are therefore difficult to apply in a medical setup, which is a major advantage of our method.

One consideration for the application of neural networks is the ability to generalize on unseen data. We tested our method on  synthetic vessel trees that were not included in the training or validation procedure.
The availability of sufficient clinical data is a common problem in medical machine learning, such that synthetically generated data is often used \cite{Li_2021_aneu}. 
However, not all flow patterns in anatomical cerebral vessel trees might be covered by synthetic cases. In particular, pathological abnormalities that alter the flow, e.g., stenoses or aneurysms, were not considered in our study. For a clinical application, our method needs to be investigated on real clinical patient data.
Our method  is not limited to predicting velocity fields for medical applications. It could be applied analogously to predict pressure distributions or other quantities of interest and is suited for any vessel or tube-shaped geometries, even outside the medical field.

\section{Conclusion}
The computational complexity of CFD simulations restricts patient-specific hemodynamics assessment in the clinical workflow. We presented a deep-learning-based CFD surrogate model tailored to predict the high-resolution spatial and temporal velocity field given a complex vascular geometry and BCs within seconds.
We envision that our approach could form the basis for a clinical hemodynamics assessment tool that supports diagnosis of vascular diseases and provides online feedback to clinicians during procedures.

\subsubsection*{Disclaimer}
The concepts and information presented are
based on research and are not commercially available.

\bibliographystyle{splncs04}
\bibliography{bib}
\end{document}